\documentclass[sigconf]{acmart}
\usepackage{tikz}
\usetikzlibrary{arrows.meta, positioning}
\usepackage{pgfplots}
\pgfplotsset{compat=1.18}
\usepackage{float}
\usepackage{amsthm}
\usepackage{amsfonts}
\usepackage{centernot}
\usepackage{graphicx}
\usepackage{siunitx}
\usepackage{multirow}
\usepackage{mathtools}
\usepackage{soul}
\usepackage{cleveref}
\usepackage{subcaption}
\usepackage{pifont}
\usepackage{algorithm}
\usepackage{algpseudocode}
\usepackage{booktabs}  % For toprule, midrule, bottomrule
\usepackage{float}     % For [H] option (if using Solution 2)
\usepackage{caption} 
\usepackage{enumitem}
\usepackage{thmtools} 
\usepackage{thm-restate}

\usepackage{booktabs}
\usepackage{makecell}
\usepackage{graphicx}
\usepackage{pifont}
\newcommand{\cmark}{\ding{51}}
\newcommand{\xmark}{\ding{55}}
\newcommand{\pmark}{$\sim$}

\declaretheorem[
    name=Theorem,
    numberwithin=section,
    refname={Theorem,Theorems},
    Refname={Theorem,Theorems}
]{theorem}

\declaretheorem[
    name=Definition,
    style=definition,
    sibling=theorem
]{definition}

\newtheorem{conjecture}{Conjecture}[section]

\usepackage{xcolor}

\begin{document}

\title{From Specification to Architecture: A Theory Compiler for
Knowledge-Guided Machine Learning}

\author{Asela Hevapathige}
\email{asela.hevapathige@unimelb.edu.au}
\affiliation{%
  \institution{AI, Optimization and Pattern Recognition Research Group \\
               Faculty of Engineering and Information Technology \\
               University of Melbourne}
  \city{Melbourne}
  \country{Australia}
}

\author{Yu Xia}
\email{yu.xia8@unimelb.edu.au}
\affiliation{%
  \institution{AI, Optimization and Pattern Recognition Research Group \\
               Faculty of Engineering and Information Technology \\
               University of Melbourne}
  \city{Melbourne}
  \country{Australia}
}

\author{Sachith Seneviratne}
\email{sachith.seneviratne@unimelb.edu.au}
\affiliation{%
  \institution{AI, Optimization and Pattern Recognition Research Group \\
               Faculty of Engineering and Information Technology \\
               University of Melbourne}
  \city{Melbourne}
  \country{Australia}
}

\author{Saman Halgamuge}
\email{saman@unimelb.edu.au}
\affiliation{%
  \institution{AI, Optimization and Pattern Recognition Research Group \\
               Faculty of Engineering and Information Technology \\
               University of Melbourne}
  \city{Melbourne}
  \country{Australia}
}

\begin{abstract}
Theory-guided machine learning has demonstrated that including authentic domain knowledge directly into model design improves performance, sample efficiency and out-of-distribution generalisation. Yet the process by which a formal domain theory is translated into architectural constraints remains entirely manual, specific to each domain formalism, and devoid of any formal correctness guarantee. This translation is non-transferable between domains, not verified, and does not scale. We propose the \emph{Theory Compiler}: a system that accepts a typed, machine-readable domain theory as input and automatically produces an architecture whose function space is provably constrained to be consistent with that theory by construction, not by regularisation. We identify three foundational open problems whose resolution defines our research agenda: (1) designing a universal theory formalisation language with decidable type-checking; (2) constructing a compositionally correct compilation algorithm from theory primitives to architectural modules; and (3) establishing soundness and completeness criteria for formal verification. We further conjecture that compiled architectures match or exceed manually-designed counterparts in generalisation performance while requiring substantially less training data, a claim we ground in classical statistical learning theory. We argue that recent advances in formal machine learning theory, large language models, and the growth of an interdisciplinary research community have made this paradigm achievable for the first time.
\end{abstract}




\maketitle

\section{Introduction}

Machine learning has become a powerful method for extracting predictive insights from data, achieving impressive results in areas such as vision, language, and decision-making by learning adaptable function approximators \cite{khan2021machine,bharadiya2023comprehensive,zheng2024survey}. This statistical approach is particularly effective in data-rich environments. However, many important scientific fields face the opposite scenario: they involve costly experiments, limited observations, and a wealth of existing formal knowledge, including physical laws, symmetry principles, and conservation constraints \cite{rosen2008symmetry,carleo2019machine}. Scientific machine learning has responded to this challenge by demonstrating that injecting domain knowledge into model design substantially improves performance, sample efficiency and out-of-distribution generalisation. The mechanism is consistent across approaches: restricting the effective hypothesis class to functions consistent with a known domain theory reduces statistical complexity. Physics-informed neural networks (PINNs)~\cite{karniadakis2021physics,cuomo2022scientific} embed differential operators directly into the loss, reducing data requirements for partial differential equation (PDE)-constrained problems. Equivariant architectures~\cite{cohen2016group,bronstein2021geometric} enforce geometric symmetries structurally and achieve state-of-the-art results in molecular property prediction. Causal representation learning~\cite{scholkopf2021toward,doehner2026causal} builds interventional invariances into generative models. What unites these is not a shared technique but a shared premise: that the right inductive bias, if correctly specified, does more work than data alone.

\begin{table*}[t]
\centering
\caption{Comparison of approaches for knowledge-guided machine learning. \cmark~= satisfied; \xmark~= not satisfied; \pmark~= partial / conditional.}
\label{tab:comparison}
\setlength{\tabcolsep}{6pt}
\renewcommand{\arraystretch}{1.25}
\begin{tabular}{lccccc}
\toprule
\textbf{Approach}
  & \textbf{\makecell{Constraints by\\Construction}}
  & \textbf{\makecell{Soundness\\Certificate}}
  & \textbf{\makecell{Domain\\Reusability}}
  & \textbf{\makecell{OOD\\Robustness}}
  & \textbf{\makecell{Zero Human\\Translation Effort}} \\
\midrule
Physics-Informed NNs (PINNs)    & \pmark & \xmark & \xmark & \pmark & \xmark \\
Equivariant Networks             & \cmark & \xmark & \xmark & \cmark & \xmark \\
Soft Constraint / Regularisation & \xmark & \xmark & \pmark & \xmark & \xmark \\
Causal Generative Models         & \pmark & \xmark & \xmark & \pmark & \xmark \\
Neural Architecture Search (NAS) & \xmark & \xmark & \pmark & \xmark & \pmark \\
AutoML                           & \xmark & \xmark & \cmark & \xmark & \cmark \\
\textbf{Theory Compiler (Ours)}  & \cmark & \cmark & \cmark & \cmark & \cmark \\
\bottomrule
\end{tabular}
\end{table*}

Despite these advances, the process by which domain knowledge is translated into architectural constraints remains entirely manual and ad hoc. A domain scientist and a machine learning engineer must collaborate to produce each encoding, requiring deep expertise in both the theory and the relevant modelling formalisms. The resulting translation does not transfer across problems and provides no formal guarantee of faithfulness to the original theory. As scientific machine learning expands to new domains such as climate modelling \cite{de2023machine,chen2023machine}, drug discovery \cite{gawehn2016deep,chen2018rise}, and materials science \cite{wei2019machine,choudhary2022recent}, this manual encoding bottleneck increasingly limits the field's reach. There is currently no formal sense in which an encoding can be said to be correct: existing architectures offer no guarantee of constraint satisfaction at inference time, and the informality of the theory-to-architecture mapping makes competing encodings impossible to compare or audit.

Existing work has focused almost entirely on how to exploit an encoded theory efficiently during training, while largely neglecting the more fundamental question of whether the encoding process itself can be automated and placed on a formal footing. This paper argues that both are achievable. We propose the \textbf{Theory Compiler}, a research paradigm to formalise the interface between domain theory and machine learning architecture and to establish the correctness guarantees that current practice entirely lacks. To this end, we formalise what it means for an architecture to be \textit{provably consistent} with a domain theory, identify three foundational open problems that define the path forward, and state a falsifiable conjecture connecting provably constrained architectures to substantial gains in sample efficiency. If this paradigm succeeds, the bottleneck separating formal scientific knowledge from deployable and trustworthy models would shift from an art practised by a few specialists to a principled engineering discipline, one capable of scaling scientific machine learning to the full breadth of domains where it is most needed.

\vspace{-3mm}

\section{Background}

\noindent\textbf{Theory-guided Machine Learning.} The idea of incorporating domain knowledge into neural architectures spans several largely independent communities. PINNs ~\cite{cuomo2022scientific,farea2024understanding}
embed differential operators into the training loss, enforcing PDE constraints softly and approximately.
Equivariant networks~\cite{cohen2016group,bronstein2021geometric}
enforce geometric symmetries structurally by design, achieving state-of-the-art results in molecular property prediction. Port-Hamiltonian networks~\cite{van2014port} and Hamiltonian neural networks~\cite{greydanus2019hamiltonian,han2021adaptable} encode energy-dissipation structure for physical systems. Causal representation learning~\cite{scholkopf2021toward,doehner2026causal}
builds interventional invariances into generative models. Interpretable neural networks for equation discovery encode algebraic structure directly into the architecture by construction \cite{ranasinghe2024ginn}. Yet each of these frameworks addresses a single formalism through manual encoding, and no existing work provides a unified compilation, nor formal correctness guarantees for the encoding process itself.

\smallskip\noindent\textbf{NAS and AutoML.} Neural Architecture Search~\cite{ren2021comprehensive,salmani2025systematic} and Automated Machine Learning ~\cite{hutter2019automated,alcobacca2026literature}
automate architecture design through search over predefined spaces, but the search space itself remains hand-crafted and unconstrained by domain theory. These methods optimise performance empirically rather than deriving structure from formal specifications, and offer no soundness guarantees with respect to domain constraints.

\begin{figure*}[t]
\centering
\begin{tikzpicture}[
  node distance=0.5cm and 0.6cm,
  box1/.style={rectangle, rounded corners=3pt, draw=blue!60, fill=blue!10,
              minimum height=0.9cm, minimum width=2.1cm, align=center,
              font=\small\bfseries, text width=2.0cm},
  box2/.style={rectangle, rounded corners=3pt, draw=teal!60, fill=teal!10,
              minimum height=0.9cm, minimum width=2.1cm, align=center,
              font=\small\bfseries, text width=2.0cm},
  box3/.style={rectangle, rounded corners=3pt, draw=orange!70, fill=orange!10,
              minimum height=0.9cm, minimum width=2.1cm, align=center,
              font=\small\bfseries, text width=2.0cm},
  box4/.style={rectangle, rounded corners=3pt, draw=violet!60, fill=violet!10,
              minimum height=0.9cm, minimum width=2.1cm, align=center,
              font=\small\bfseries, text width=2.0cm},
  box5/.style={rectangle, rounded corners=3pt, draw=green!60!black, fill=green!10,
              minimum height=0.9cm, minimum width=2.1cm, align=center,
              font=\small\bfseries, text width=2.0cm},
  subbox1/.style={rectangle, rounded corners=2pt, draw=blue!30, fill=blue!5,
              minimum height=0.6cm, minimum width=2.1cm, align=center,
              font=\footnotesize, text width=2.0cm},
  subbox2/.style={rectangle, rounded corners=2pt, draw=teal!30, fill=teal!5,
              minimum height=0.6cm, minimum width=2.1cm, align=center,
              font=\footnotesize, text width=2.0cm},
  subbox3/.style={rectangle, rounded corners=2pt, draw=orange!40, fill=orange!5,
              minimum height=0.6cm, minimum width=2.1cm, align=center,
              font=\footnotesize, text width=2.0cm},
  subbox4/.style={rectangle, rounded corners=2pt, draw=violet!30, fill=violet!5,
              minimum height=0.6cm, minimum width=2.1cm, align=center,
              font=\footnotesize, text width=2.0cm},
  subbox5/.style={rectangle, rounded corners=2pt, draw=green!40!black, fill=green!5,
              minimum height=0.6cm, minimum width=2.1cm, align=center,
              font=\footnotesize, text width=2.0cm},
  arr/.style={-{Stealth[length=5pt]}, thick, gray!70}
]

\node[box1] (T)  {\textbf{Domain}\\\textbf{Theory}};
\node[box2, right=of T]  (P)  {\textbf{Theory}\\\textbf{Parser}};
\node[box3, right=of P]  (IR) {\textbf{Typed}\\\textbf{Primitives}};
\node[box4, right=of IR] (S)  {\textbf{Architecture}\\\textbf{Synthesizer}};
\node[box5, right=of S]  (M)  {\textbf{Compiled}\\\textbf{Model}};

\node[subbox1, below=0.3cm of T]  {Symmetries, differential operators, conservation laws, causal graphs};
\node[subbox2, below=0.3cm of P]  {Type inference and well-formedness checking};
\node[subbox3, below=0.3cm of IR] {Universal intermediate representation};
\node[subbox4, below=0.3cm of S]  {Compilation rules and compositional correctness};
\node[subbox5, below=0.3cm of M]  {Architecture, loss function, soundness certificate};

\draw[arr] (T)  -- (P);
\draw[arr] (P)  -- (IR);
\draw[arr] (IR) -- (S);
\draw[arr] (S)  -- (M);

\end{tikzpicture}
\caption{The Theory Compiler pipeline. A domain theory is parsed into typed primitives within a universal intermediate representation. The Architecture Synthesizer applies formal compilation rules to map each primitive to a canonical architectural module and composes them according to the relational structure of the theory. The output is a compiled model accompanied by a soundness certificate verifying consistency with the input theory.}
\label{fig:pipeline}
\end{figure*}
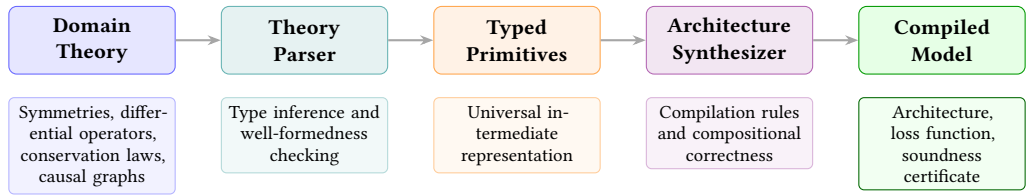

\smallskip\noindent\textbf{Formal Foundations and Compilation.}
Program synthesis~\cite{gulwani2017program} and verified
compilation~\cite{leroy2009formal} methods show that formal specifications can be automatically translated into correct artefacts, but both target executable code rather than neural architectures. Recent category-theoretic work~\cite{bronstein2021geometric,cruttwell2022categorical}
provides the algebraic language needed to reason about architectural constraints formally, and formal verification has been applied to small neural network properties~\cite{urban2020perfectly}, showing that machine-checkable certificates for architectures are feasible in principle. Finally, language models trained on scientific
text~\cite{taylor2022galactica,lewkowycz2022solving} can parse mathematical structure automatically, making the theory-parsing step of our pipeline practically achievable.

Compared to existing methods, our proposed approach offers two key properties that current methods lack. First, by provably restricting the hypothesis class to functions consistent with the domain theory, they address data scarcity. Second, the compilation rules and soundness certificates are explicit artefacts, providing a degree of accountability that post-hoc explainability methods cannot match. Table~\ref{tab:comparison} summarises how existing approaches compare across diverse properties that the Theory Compiler is designed to satisfy.

\section{A Timely Confluence}
\label{sec:convergence}

The Theory Compiler unifies ideas that have emerged independently 
across several communities: symmetry has been linked to equivariant 
architectures, differential equations to physics-informed networks, 
and causal structure to generative models. What has been overlooked 
is that these are all instances of the same underlying problem: 
translating a formal domain theory into a constrained architecture. 
Recent advances in algebraic machine learning theory, scientific 
language models, and formal verification have converged to make a 
unified solution achievable for the first time.

\smallskip\noindent\textbf{The mathematical foundations are in place.} Recent work in machine learning theory has begun to describe neural network operations in precise algebraic terms. Symmetry-preserving layers can now be characterised as natural transformations between structured spaces~\cite{bronstein2021geometric}, and differential operators admit clean algebraic representations~\cite{cruttwell2022categorical}. This growing body of work suggests that a common formal language capable of expressing symmetries, differential equations, and causal constraints in a unified way is within reach.

\smallskip\noindent\textbf{Theory parsing is becoming tractable.} Extracting formal structure from domain knowledge has become far more tractable through a combination of structured specification languages and language models trained on scientific text~\cite{taylor2022galactica}. We do not claim that current models can parse arbitrary theory from natural language; rather, when theories are expressed in a lightweight domain-specific language, LLM assistance can automate the labour-intensive alignment between informal scientific prose and typed formal primitives. This opens a realistic path to reducing the manual effort that the parsing step currently requires in each new domain.

\smallskip\noindent\textbf{The right community is already here.} Building the Theory Compiler requires machine learning theorists, programming language researchers, domain scientists, and engineers working together. Conferences such as KDD span all of these groups, and the application domains where this work matters most, including drug discovery, climate modelling, and materials science, are already well represented in these communities.

\section{Theory Compiler} \label{sec:pipeline}

In this section, we propose the \textbf{Theory Compiler}, a system that accepts a formal domain theory as input and automatically produces an architecture, a structured loss, and an evaluation protocol whose inductive biases are derived from that theory by construction, not by approximation.The proposed pipeline proceeds in three stages:

\begin{itemize}[leftmargin=1.2em,itemsep=1pt,topsep=2pt]
  \item \textbf{Theory Parser.} Reads the domain theory and extracts its core components, such as symmetry groups, differential operators, conservation quantities, and causal graphs, each assigned a type in a universal intermediate representation.
  \item \textbf{Architecture Synthesizer.} Maps each typed component to a canonical architectural module and composes these modules according to the relational structure of the theory.
  \item \textbf{Compiled Model.} The resulting architecture, structured loss, and evaluation protocol form a complete model specification whose design is fully determined by, and provably consistent with, the input theory.
\end{itemize}

Intuitively, a domain theory is a formal specification of the structural constraints a model must satisfy, independently of the data it is trained on. We make this precise as follows.

\begin{definition}[Domain Theory]
A domain theory $\mathcal{T} = (\Sigma, \Phi, \mathcal{R})$ consists of a typed signature $\Sigma$ representing the quantities of interest, a set of formal primitives $\Phi$ comprising symmetry groups, differential operators, conservation functionals, and causal graphs, and a set of relations $\mathcal{R}$ governing their composition.
\end{definition}

A compiled model should not merely approximate the constraints expressed in $\mathcal{T}$, but satisfy them exactly. This motivates the following notions of soundness and completeness.

\begin{definition}[Soundness]
A compiled model is \emph{sound} with respect to a domain theory $\mathcal{T}$ if every output satisfies all constraints in $\mathcal{R}$ for all inputs, not merely in expectation over a training distribution.
\end{definition}

\begin{definition}[Completeness]
A compiled model is \emph{complete} with respect to a domain theory
$\mathcal{T}$ if every hypothesis consistent with $\mathcal{T}$ is
realisable by the architecture, i.e.\ no valid inductive bias
expressible under $\mathcal{T}$ is excluded by the compilation
process.
\end{definition}

Soundness and completeness together ensure that the compiled
architecture's function space is in
exact correspondence with the theory-consistent hypothesis class,
independently of training data.

\begin{figure*}[H]
\centering
\scalebox{1.0}{
\begin{tikzpicture}[
  node distance=0.5cm and 0.6cm,
  box/.style={rectangle, rounded corners=3pt, draw=black!60, fill=gray!7,
              minimum height=0.9cm, minimum width=2.1cm, align=center,
              font=\small, text width=2.0cm},
  subbox/.style={rectangle, rounded corners=2pt, draw=black!30, fill=white,
              minimum height=0.6cm, minimum width=2.1cm, align=center,
              font=\footnotesize, text width=2.0cm},
  arr/.style={-{Stealth[length=5pt]}, thick, black!50}
]
\node[box] (T) {\textbf{Domain}\\\textbf{Theory}};
\node[box, right=of T] (P) {\textbf{Theory}\\\textbf{Parser}};
\node[box, right=of P] (IR) {\textbf{Typed}\\\textbf{Primitives}};
\node[box, right=of IR] (S) {\textbf{Architecture}\\\textbf{Synthesizer}};
\node[box, right=of S] (M) {\textbf{Compiled}\\\textbf{Model}};

\node[subbox, below=0.3cm of T]  {Symmetries, differential operators, conservation laws, causal graphs};
\node[subbox, below=0.3cm of P]  {Type inference and well-formedness checking};
\node[subbox, below=0.3cm of IR] {Universal intermediate representation $\mathcal{L}_\mathcal{T}$};
\node[subbox, below=0.3cm of S]  {Compilation rules and compositional correctness};
\node[subbox, below=0.3cm of M]  {Architecture, loss function, soundness certificate};

\draw[arr] (T)  -- (P);
\draw[arr] (P)  -- (IR);
\draw[arr] (IR) -- (S);
\draw[arr] (S)  -- (M);
\end{tikzpicture}
}
\caption{The Theory Compiler pipeline. A domain theory is parsed into typed primitives within a universal intermediate representation. The Architecture Synthesizer applies formal compilation rules to map each primitive to a canonical architectural module and composes them according to the relational structure of the theory. The output is a compiled model accompanied by a soundness certificate verifying consistency with the input theory.}
\label{fig:pipeline}
\end{figure*}

\subsection{Theory Parser}
The Theory Parser reads a formal domain theory $\mathcal{T} = (\Sigma, \Phi, \mathcal{R})$ and extracts its core components into a typed intermediate representation $\mathcal{L}_\mathcal{T}$. Each primitive $\phi \in \Phi$ is assigned a type from a fixed type system: a symmetry group $G$ receives type \texttt{Sym}, a differential operator $\mathcal{L}$ receives type \texttt{Diff}, a conservation functional $\mathcal{C}$ receives type \texttt{Cons}, and a causal graph $\mathcal{G}$ receives type \texttt{Caus}. Well-formedness of $\mathcal{T}$ is verified at this stage by checking that all relations in $\mathcal{R}$ are type-consistent, ensuring that the subsequent synthesis stage operates on a valid specification.

\subsection{Architecture Synthesizer}
The Architecture Synthesizer applies a compilation map $\kappa: \Phi \to \mathcal{M}$ that assigns each typed primitive a canonical architectural module in a model space $\mathcal{M}$. For a primitive $\phi \in \Phi$ with type $\tau \in \mathcal{L}_\mathcal{T}$, the general form is
\begin{equation}
  \kappa(\phi : \tau) \;\mapsto\; \mathcal{M}_\tau(\phi),
\end{equation}
where $\mathcal{M}_\tau(\phi)$ denotes the canonical architectural module associated with type $\tau$ and parameterised by $\phi$. The mapping is extensible; new theory types can be accommodated by registering additional compilation rules. Table~\ref{tab:rules} summarises the representative compilation rules; the framework is extensible and any primitive for which a sound architectural realisation can be defined admits an additional rule.

\begin{table}[h]
\centering
\caption{Representative compilation rules $\kappa$ mapping typed primitives to canonical architectural modules.}
\label{tab:rules}
\small
\begin{tabular}{lll}
\toprule
\textbf{Type} & \textbf{Primitive} $\phi$ & \textbf{Module} $\mathcal{M}_\tau(\phi)$ \\
\midrule
\texttt{Sym}  & Symmetry group      & Equivariant layer \\
\texttt{Cons} & Conservation law    & Hard constraint module \\
\texttt{Diff} & Differential operator & Constrained function class \\
\texttt{Caus} & Causal graph        & Structured generative model \\
\vdots        & \vdots              & \vdots \\
\bottomrule
\end{tabular}
\end{table}

 When $\mathcal{T}$ contains multiple interacting primitives, the synthesizer composes their modules according to $\mathcal{R}$, with the formal requirement that $\kappa$ extends to a functor from the category of theories to the category of model spaces, preserving constraint composition across all interactions specified in the theory.

\subsection{Compiled Model with Correctness}
The output of the pipeline is a compiled model $\mathcal{A}(\mathcal{T})$ together with a soundness certificate. Defining the theory-consistent hypothesis class as
\begin{equation}
  \mathcal{H}(\mathcal{T}) = \{f : f \models \mathcal{T}\},
\end{equation}
the compiled model satisfies two formal properties. Soundness requires
\begin{equation}
  \mathcal{A}(\mathcal{T}) \subseteq \mathcal{H}(\mathcal{T}),
\end{equation}
guaranteeing that no function realisable by the architecture violates the domain theory. Completeness requires
\begin{equation}
  \mathcal{H}(\mathcal{T}) \subseteq \mathcal{A}(\mathcal{T}),
\end{equation}
ensuring that no valid hypothesis is discarded by the compilation process. Together, these properties establish that the compiled model's function space is in exact correspondence with the set of functions the theory permits, independently of the training data.

\section{Challenges, Open Problems, and a Conjecture}
\label{sec:challenges}

We now formalise the core technical barriers that the Theory Compiler must overcome, present three foundational open problems that define the research agenda, and state a falsifiable conjecture on the expected gains from provably constrained architectures.

\begin{table*}[h]
\centering
\caption{Research milestones for the Theory Compiler paradigm.}
\label{tab:milestones}
\small
\setlength{\tabcolsep}{4pt}
\begin{tabular}{p{1.4cm} p{4.2cm} p{4.2cm} p{4.2cm}}
\toprule
\textbf{Horizon} & \textbf{Target} & \textbf{Key Challenge} 
  & \textbf{Success Criterion} \\
\midrule
Near term \newline (1--2 yrs)
  & Single-domain compiler for classical mechanics or molecular symmetries
  & Decidable type-checking for $\mathcal{L}_\mathcal{T}$
  & Known architecture derived automatically with a machine-checkable 
    soundness proof \\
\addlinespace
Medium term \newline (3--5 yrs)
  & Multi-primitive compiler with LLM-assisted theory parsing
  & Compositional correctness across interacting primitives
  & Novel inductive bias synthesised with measured gain in the low-data regime \\
\addlinespace
Long term \newline (5$+$ yrs)
  & Full pipeline from theory specification to verified architecture
  & Certificates for nonlinearly interacting constraints
  & Reliable OOD extrapolation across diverse domains \\
\bottomrule
\end{tabular}
\end{table*}

\subsection{Implicit Assumptions in Current Practice}

Current approaches to theory-guided machine learning rest on three assumptions that are rarely questioned. The Theory Compiler challenges each of them.

\paragraph{\textbf{Assumption 1: Theory encoding is inherently manual.}}
Scientific knowledge is currently embedded into neural architectures through one-off design decisions made by human experts, a process that cannot be reused across problems or formally verified. The Theory Compiler treats this encoding as a well-defined computation: given a formal theory, a compiler should produce an architecture, and that transformation should be checkable for correctness just as any program transformation is.

\paragraph{\textbf{Assumption 2: Soft constraint enforcement is the practical default.}}
Penalty-based methods dominate largely for practical reasons: they are easy to implement, differentiable, and compatible with gradient-based optimisation. But a model trained with soft constraints may satisfy them on the training distribution while violating them under distribution shift, since the architecture itself imposes no structural restriction. On the other hand, hard constraint methods define models that can’t break the rules at all, making it clear whether the constraints are being followed or not \cite{bahavan2025sphor}. 

\paragraph{\textbf{Assumption 3: Theory-guided methods must be formalism-specific.}}
Current frameworks each target a single formalism — equivariant networks for symmetry groups, PINNs for differential equations, Port-Hamiltonian networks for energy dissipation. These communities have developed largely in parallel, with their own notation and techniques. The Theory Compiler unifies them to a single formal framework.

\subsection{Open Problems} \label{sec:barriers}

We state three foundational open problems that define the Theory Compiler research agenda.

\paragraph{\textbf{Problem I: The theory language.}}
A theory specification language must be expressive enough to capture the full range of domain constraints in scientific ML, yet structured enough to admit automatic well-formedness checking. Rich type systems can represent the necessary constraints but lose decidability; simpler ones remain checkable but cannot express many physically meaningful theories.A layered design, consisting of a decidable core extended by a richer expressive layer, seems most promising, but where to draw the boundary and how to compose layers cleanly remains an open problem at the intersection of programming language theory and machine learning.

\paragraph{\textbf{Problem II: Compositional correctness.}}
Individual theory-to-architecture mappings are well established: symmetry groups compile to equivariant layers, conservation laws to linear constraints, causal graphs to structured generative models. The difficulty arises when a theory contains multiple interacting primitives---naively stacking modules can satisfy each constraint in isolation while violating their joint structure. Formally, letting $\kappa$ denote the compilation map, the conjunction of theories should satisfy $\kappa(\mathcal{T}_1 \land \mathcal{T}_2) = \kappa(\mathcal{T}_1) \otimes \kappa(\mathcal{T}_2)$ for some composition operation $\otimes$ on architecture families. Whether such a map exists, under what conditions it is unique, and what structure the specification language must impose to guarantee it, are open questions with no existing solution.

\paragraph{\textbf{Problem III: Soundness and completeness certificates.}}
A compiled architecture must satisfy soundness ($\mathcal{A}(\mathcal{T}) \subseteq \mathcal{H}(\mathcal{T})$), ruling out theory-violating functions, and completeness ($\mathcal{H}(\mathcal{T}) \subseteq \mathcal{A}(\mathcal{T})$), ensuring no valid hypotheses are discarded. For simple algebraic constraints such as equivariance, existing 
theory already tells us how to build sound and complete architectures. 
For nonlinear differential constraints interacting with conservation 
laws, no equivalent theory exists. The goal is to produce 
machine-checkable proofs of both properties for any compiled 
architecture, along with useful diagnostics when a theory does not 
uniquely determine the architecture. How hard this verification 
problem is remains open.

\subsection{A Falsifiable Conjecture}

A framework that constrains architectures should provide measurable practical benefits, which we propose as a falsifiable conjecture.

\begin{conjecture}[Theory Compiler Conjecture]
\label{conj:main}
For any domain admitting a formalized theory $\mathcal{T} \in \mathcal{L}_\mathcal{T}$, and any training set $S$ drawn from a distribution consistent with $\mathcal{T}$, the compiled architecture $\mathcal{A}(\mathcal{T})$ achieves generalisation error no worse than any manually-designed architecture $\mathcal{A}^*$ encoding $\mathcal{T}$ approximately, while requiring asymptotically fewer training examples.
\end{conjecture}

The intuition follows from standard learning theory. When $\mathcal{A}^*$ enforces constraints softly, its hypothesis class $\mathcal{H}(\mathcal{A}^*)$ contains functions that satisfy constraints only approximately or only on the training support. In contrast, $\mathcal{H}(\mathcal{A}(\mathcal{T}))$ contains only functions satisfying constraints exactly everywhere. The strict inclusion $\mathcal{H}(\mathcal{A}(\mathcal{T})) \subset \mathcal{H}(\mathcal{A}^*)$ then yields, via classical complexity bounds, a sample complexity advantage for the compiled architecture \cite{blumer1989learnability}. Three testable sub-claims sharpen the conjecture:

\begin{enumerate}
    \item \textbf{Covariate shift robustness.} On inputs outside the training support, compiled architectures degrade more slowly in constraint satisfaction than manually-designed counterparts with soft constraints.
    
    \item \textbf{Meta-learning the compilation map.} The map $\kappa$ is itself learnable: a meta-model trained on theory-architecture pairs from known domains can propose plausible architectures for novel theories, reducing the manual effort of extending $\mathcal{L}_\mathcal{T}$.
    
    \item \textbf{Empirical superiority across domains.} Across benchmark problems in chemistry, physics, and biology, hard-constraint compilation strictly dominates soft-penalty baselines in the low-data regime.
\end{enumerate}

The conjecture is falsifiable. If a domain with a clean formal theory shows no benefit from compilation, or if soft constraints are shown to match hard constraints asymptotically with enough data, the conjecture fails. The advantage likely varies across theory classes, and understanding where and why is itself an important open question.

\section{Illustrative Example: DEE Gene Prioritization}
\label{sec:example}

Gene prioritization for Developmental and Epileptic Encephalopathy 
(DEE) \cite{scheffer2024developmental} aims to rank candidate genes 
by their likelihood of causing the disorder. Confirmed causal genes 
are scarce, regulatory mechanisms involve multiple interacting 
biological constraints, and encoding these constraints currently 
requires significant domain expertise without formal correctness 
guarantees \cite{riva2025pathophysiological}.

We represent $\mathcal{T}_{\text{DEE}}$ using three primitives: a 
causal graph $\phi_G$ over gene regulatory relationships 
($\mathsf{Caus}$), a conservation constraint $\phi_C$ capturing 
pathway flux conservation ($\mathsf{Cons}$), and a permutation 
symmetry $\phi_K$ over genes within shared pathways ($\mathsf{Sym}$). 
The Theory Parser verifies type-consistency of relations in $R$ and 
emits a typed specification in $\mathcal{L}_\mathcal{T}$. The 
Architecture Synthesizer then compiles each primitive to a canonical 
module: $\phi_G$ to a direction-aware GNN, $\phi_C$ to a hard 
conservation constraint layer, and $\phi_K$ to an equivariant 
aggregation. The resulting architecture provably cannot violate 
regulatory directionality or pathway symmetry by construction, 
without any penalty term in the loss.

This example surfaces all three open problems: whether the type system 
can express interacting biological constraints with decidable 
well-formedness (Problem~I); whether correct compilation requires 
deriving $\kappa(\phi_G \wedge \phi_K)$ rather than composing 
$\kappa(\phi_G)$ and $\kappa(\phi_K)$ independently (Problem~II); 
and whether a machine-checkable soundness certificate can be produced 
for the resulting architecture (Problem~III). Resolving all three in 
this setting is a central goal of the research agenda.

\section{Milestones Toward a Certified Science}

Table~\ref{tab:milestones} summarises the expected progression of the research paradigm across three time horizons. The near-term goal is not to produce architectures that are not already known, but to demonstrate that their derivation can be automated with formal correctness guarantees. The medium-term goal is to move beyond single-domain compilation toward compositional correctness across interacting theory primitives. The long-term endpoint is a domain scientist who specifies governing equations and symmetries in $\mathcal{L}_\mathcal{T}$, receives a certified architecture in return, and trains it on a small dataset to obtain a model that performs reliably beyond its training distribution. 

\bibliography{references}
\bibliographystyle{ACM-Reference-Format}

\end{document}